\documentclass{elsarticle}

\journal{}

\usepackage[utf8]{inputenc} 	
\usepackage{graphicx}      		
\usepackage[numbers]{natbib}	
\usepackage{amsfonts}        	
\usepackage{amsmath,amssymb}    
\usepackage{bm}					
\usepackage{xcolor}             
\usepackage{float}              
\usepackage[caption=false]{subfig} 
\usepackage{multirow}  			
\usepackage{tikz}				

\newcommand{\figasset}{
\begin{tikzpicture}

\def\MAX{7.0}
\def\UNIT{0.1*\MAX}
\def\HALF{0.5*\UNIT}
\def\QUART{0.5*\HALF}

\def\VA{\HALF}
\def\VB{\VA + \HALF}
\def\VC{\VA + \UNIT}
\def\VW{0.2*\UNIT}

\def\YPIPE{2.5*\UNIT}

\def\XSEP{7*\UNIT}
\def\SW{1.0*\UNIT}
\def\SR{0.5*\UNIT}
\def\XSB{\XSEP+\SR}
\def\XSC{\XSB+\SW}
\def\XSD{\XSC+\SR}
\def\XSM{\XSB+0.5*\SW}
\def\XQ{\XSD+\HALF}

\def\YQW{\YPIPE-\SR-\QUART}
\def\YQG{\YPIPE+\SR+\QUART}

\foreach \x/\l in {1, 2, 3, 5/J}{

    \draw [thick, dashed] (\UNIT*\x, 0) -- (\UNIT*\x, -\HALF);
    \draw [thick, ] (\UNIT*\x, 0) -- (\UNIT*\x, \VA);
    \draw [thick, ->] (\UNIT*\x, \VC) -- (\UNIT*\x, \YPIPE);
    \draw [thick, ->] (\UNIT*\x-\VW, \VA) -- (\UNIT*\x+\VW, \VA) -- (\UNIT*\x-\VW, \VC) -- (\UNIT*\x+\VW, \VC) -- cycle;
    
    \node[anchor=west] at (\UNIT*\x, \VB) {$\l$};
}
\draw [thick, dotted] (\UNIT*3.5, \VB) -- (\UNIT*4.5, \VB);

\draw [thick, ->] (\UNIT, \YPIPE) -- (\XSEP, \YPIPE);

\draw [thick] (\XSB, \YPIPE-\SR) arc (270:90:\SR cm);
\draw [thick] (\XSB, \YPIPE-\SR) -- (\XSC, \YPIPE-\SR);
\draw [thick] (\XSB, \YPIPE+\SR) -- (\XSC, \YPIPE+\SR);
\draw [thick] (\XSC, \YPIPE-\SR) arc (-90:90:\SR cm);

\draw [thick] (\XSC, \YPIPE-\SR) -- (\XSC, \YPIPE+0.25*\SR); 

\draw [thick, ->] (\XSD, \YPIPE) -- (\XQ, \YPIPE) node[anchor=west]{Oil};
\draw [thick, ->] (\XSM, \YPIPE+\SR) -- (\XSM, \YQG) -- (\XQ, \YQG) node[anchor=west]{Gas}; 
\draw [thick, ->] (\XSM, \YPIPE-\SR) -- (\XSM, \YQW) -- (\XQ, \YQW) node[anchor=west]{Water}; 

\end{tikzpicture}
}

\newcommand{\figwelltopology}{
	\begin{tikzpicture}
	
	\def\MAX{5.5}
	\def\HALF{0.05*\MAX}
	\def\UNIT{0.1*\MAX}
	\def\XA{0.2*\MAX}
	\def\XB{0.4*\MAX}
	\def\MID{0.5*\MAX}
	\def\XC{0.6*\MAX}
	\def\XD{0.8*\MAX}
	
	\foreach \x/\l in {\XA / {Upstream}, \XD/ {Downstream}}{
		\draw[fill] (\x, 0) circle (0.05cm);
	}
	
	\draw[thick] (\XB,-\HALF) -- (\XB,\HALF) -- (\XC,-\HALF) -- (\XC,\HALF) -- cycle;
	
	\draw [thick, ->] (0, 0) -- (\XB, 0);
	\draw [thick, ->] (\XC, 0) -- (\MAX, 0);
	
	\foreach \x/\l in {\XA / {$p_1, T$}, \MID/ {$u$}, \XD / {$p_2$}}{
		\draw [thick,dashed] (\x, 0) -- (\x, \UNIT);
		\draw[thick] (\x, \UNIT+0.5) circle (0.5cm);
		\node[anchor=center] at (\x, \UNIT+0.5) {\l};
	}
	
	\draw [thick, ->] (\XB, -1) -- (\XC, -1);
	\node[anchor=north] at (\MID, -\HALF) {$q$};
	
	\end{tikzpicture}
}

\newcommand{\figarchitecture}{
\begin{tikzpicture}

\def\GY{5}
\def\GA{1}
\def\GB{\GA+2}
\def\HA{4}
\def\HB{\HA+2}

\node[anchor=east] at (0, 2) {$x_{ij}$};
\draw [thick, ->] (0, 2) -- (\GA, 2);

\node[anchor=east] at (0, 1) {$\beta_{j}$};
\draw [thick, ->] (0, 1) -- (\HA, 1);

\draw[thick, fill=white] (\GA, 1.5) -- (\GB, 1.5) -- (\GB, 2.5) -- (\GA, 2.5) -- cycle;
\node[anchor=center] at (\GA+1.0, 2) {$g(x_{ij}; \gamma_j)$};
\draw [thick, ->] (\GB, 2) -- (\HA, 2);
\node[anchor=south] at (\GB+0.5, 2) {$z_{ij}$};

\draw[thick, fill=white] (\HA, 0.5) -- (\HB,0.5) -- (\HB,2.5) -- (\HA,2.5) -- cycle;
\node[anchor=center] at (\HA+1, 1.5) {$h(z_{ij}; \beta_j, \alpha)$};

\draw [thick, ->] (\HB, 1.5) -- (\HB+1, 1.5) node[anchor=west] {$Q_{ij}$};
\end{tikzpicture}
}

\newcommand{\figrnn}{
\begin{tikzpicture}

\def\W{0.5}
\def\D{0.25}
\def\R{0.125}

\node[anchor=east] at (0, 0) {$z_{ij}^{k}$};

\foreach \x/\l in {1/A, 2/L, 3.5/A, 4.5/L}{
    \draw[thick, ] (\x,-\D) -- (\x+\W,-\D) -- (\x+\W,\D) -- (\x,\D) -- cycle;
    \node[anchor=center] at (\x+\D, 0) {\l};
}

\draw [thick, ->] (0, 0)  -- (1, 0);
\draw [thick, ->] (1.5, 0)  -- (2, 0);
\draw [thick, ->] (2.5, 0)  -- (3.5, 0)node[midway, above]{$z_{ij}^{k+1}$};
\draw [thick, ->] (4, 0)  -- (4.5, 0);
\draw [thick, ->] (5, 0)  -- (5.5-\R, 0);
\draw [thick, ->] (5.5+\R, 0)  -- (6, 0);

\node[anchor=west] at (6, 0) {$z_{ij}^{k+2}$};

\draw[thick, fill=white](5.5, 0) circle (\R);
\node[anchor=center] at (5.5, 0) {$+$};
\draw[fill] (0.5, 0) circle (0.05cm);
\draw[thick, ->] (0.5, 0) -- (0.5, -\W) -- (5.5, -\W) -- (5.5, -\R);

\end{tikzpicture}
}

\begin{document}

\begin{frontmatter}
\title{Multi-task learning for virtual flow metering}

\author[ssaddress,uioaddress]{Anders T. Sandnes\corref{mycorrespondingauthor}}
\cortext[mycorrespondingauthor]{Corresponding author}
\ead{anders.t.sandnes@gmail.com}

\author[ssaddress,ntnuaddress]{Bjarne Grimstad}
\author[uioaddress]{Odd Kolbjørnsen}

\address[ssaddress]{Solution Seeker AS, Oslo, Norway}
\address[uioaddress]{Department of Mathematics, University of Oslo, Oslo, Norway}
\address[ntnuaddress]{Department of Engineering Cybernetics, Norwegian University of Science and Technology, Trondheim, Norway}

\begin{abstract}
Virtual flow metering (VFM) is a cost-effective and non-intrusive technology for inferring multiphase flow rates in petroleum assets. Inferences about flow rates are fundamental to decision support systems that operators extensively rely on. Data-driven VFM, where mechanistic models are replaced with machine learning models, has recently gained attention due to its promise of lower maintenance costs. While excellent performances in small sample studies have been reported in the literature, there is still considerable doubt about the robustness of data-driven VFM. In this paper, we propose a new multi-task learning (MTL) architecture for data-driven VFM. Our method differs from previous methods in that it enables learning across oil and gas wells. We study the method by modeling 55 wells from four petroleum assets and compare the results with two single-task baseline models. Our findings show that MTL improves robustness over single-task methods, without sacrificing performance.
MTL yields a 25-50\% error reduction on average for the assets where single-task architectures are struggling.
\end{abstract}


\end{frontmatter}


\section{Introduction} \label{sec:introduction}
Knowledge of gas, oil, and water flow rates in a petroleum asset is highly valuable in operations and production planning but challenging to obtain \citep{Hansen2019}. There is a large economic incentive for operators to maintain high production rates and avoid operational problems. 
Flow rates from individual wells support many important operational decisions, 
such as production optimization \citep{Foss2018}, reservoir management \citep{Kanshio2020}, and flow assurance \citep{Jamaluddin2012}. 

\begin{figure}[ht]
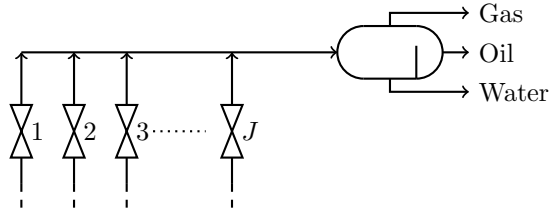

    \centering
    \figasset
    \caption{Asset with $J$ wells sharing a single separator.}
    \label{fig:asset}
\end{figure}

Most assets consist of a set of wells that produce to a shared processing facility, as illustrated in Figure \ref{fig:asset}.
The joint flow from all wells is continuously measured after being physically separated into its main phases, gas, oil, and water. These can be accurately measured by single phase flow sensors. Flow rates from individual wells are conventionally measured by routing the flow to a dedicated test separator.
The resulting observations, known as \emph{well tests}, are of high quality \citep{nfogm2005}. However, the frequency of well tests is low since the test separator accommodates one well at a time and requires several hours to measure the flow. It is therefore desirable to measure well flow rates before separation.

There are two main strategies for measuring multiphase flow, multiphase flow meters (MPFM) and virtual flow meters (VFM) \citep{Bikmukhametov2020}.
MPFMs are complex and expensive measurement devices physically installed in the well. VFM is a soft sensing technology that makes inferences about flow rates from existing sensor data and mathematical models implemented in software. VFM is often seen as complementary to MPFMs. 
Multiphase flow measurements have higher uncertainty than single phase measurements. 
Single phase measurements have errors around 0.25\% for oil and 1\% for gas rates \citep{thorn2012}. The quality and availability of water measurements are more varying.
A calibrated MPFM is expected to have approximately 5\% error for all phases\citep{thorn2012}. 
However, they are specialized to certain operating conditions and must be re-calibrated as conditions change\citep{nfogm2005}.

VFMs can be categorized based on their use of mechanistic or data-driven models \citep{Bikmukhametov2020}. A \emph{mechanistic VFM} is derived from first principles and utilizes empirical correlations sparingly. 
A \emph{data-driven VFM} is based on a machine learning method that fits a generic mathematical model to data. The generic models do not offer a physical interpretation of the parameters, as opposed to mechanistic models where parameters are related to physical properties.  
Most VFM solutions today are based on mechanistic models implemented in multiphase flow simulators \citep{Amin2015}. There are few, if any, commercially available data-driven VFM solutions. However, there has been an increasing interest in their development \citep{Bikmukhametov2020}, which is likely motivated by several factors. First, both instrumentation and data availability have improved. Second, the tooling for machine learning has improved considerably and the number of practitioners has increased. Third, oil and gas profit margins have decreased, leading to a search for more cost-efficient solutions. 
Data-driven VFM is attractive in terms of cost efficiency due to the promise of low maintenance requirements and high scalability. Data-driven VFMs are expected to be easier to develop and maintain since they only require flow rate observations to be calibrated \citep{AL-Qutami2018}. This is in contrast to mechanistic models, which can be challenging to maintain due to high model complexity \citep{Stenhouse2008}. Calibration demands flow rate observations and experiment data, such as fluid samples, to attain physically meaningful parameters values. Furthermore, calibration often requires personnel with asset experience and expert knowledge of multiphase flow physics and the VFM software. 

A diverse set of methods, models, and experiment setups for data-driven VFM have been presented in the literature.
The recent survey in \citep{Bikmukhametov2020} tabulates a selection of the proposed solutions,
where architectures based on neural networks are the most frequent.
Neural networks have been researched extensively and have been successfully applied in other domains,
such as image analysis \citep{yu2019, hong2015}, medicine \citep{hannun2019}, and natural language processing \citep{vaswani2017},
which motivates its popularity in VFM applications.
Representative works on neural network based VFMs include \citep{AL-Qutami2017b, AL-Qutami2018}, which report 2.2--4.2\% errors and 2.4--4.7\% errors respectively.
A hybrid solution of neural networks and regression trees is presented in \citep{Al-Qutami2017a}, reporting errors in the range of 1.5--6.5\%.
Gradient boosted trees are explored as an alternative to neural networks in \citep{Bikmukhametov2019}, achieving errors of 2--6\% in different scenarios.
In all approaches, the VFM model is trained on data from a single well.

Several of the proposed solutions rival the expected performance of conventional MPFMs, but commercially viable alternatives have yet to emerge.
The recent study in \citep{Grimstad2021} applied Bayesian neural networks to data-driven VFM. The authors questioned if a robust data-driven method can be obtained by individually modeling wells from historical observations. Several challenges facing any data-driven VFM were highlighted. To reiterate, there are usually few data points for each individual well, making it difficult to identify complex models. Additionally, the underlying process is non-stationary, which makes past data less relevant for future predictions. Finally, the operational practices on most assets may result in low data variety and create highly correlated explanatory variables.

Challenges related to insufficient data are common in machine learning. 
A solution is to utilize data collected from other related problems\citep{Lu2015, zhang2021}. 
There are several ways such data could be combined.
One approach is Multi-Task Learning (MTL), where models for all problems are jointly optimized \citep{zhang2021,Goodfellow-et-al-2016}. 
In MTL, the problem is given as a set of tasks, $\lbrace \mathcal{T}_1, \dots, \mathcal{T}_J\rbrace$,
where each task $\mathcal{T}_j$ has a set of observations $(y_{ij}, x_{ij}), i = 1, \dots, N_j$.
MTL attempts to jointly learn models for each task, utilizing the knowledge from other tasks to improve performance.
Tasks are assumed to share some common structure that enables the transfer of knowledge.
Several mechanisms have been suggested to facilitate knowledge sharing. 
One approach is to have a set of parameters shared between the task models. 

Methods that combine MTL and deep learning have been successfully applied to several domains, e.g., image analysis \citep{hong2019}, natural language processing \citep{sigtia2020}, and speech processing \citep{majumder2019}. It has also been applied to problems in the energy sector, such as solar and wind power \citep{wu2021, Dorado-Moreno2020}.
Multi-task neural networks have been presented in a wide range of complexities,
from simple feed forward networks \citep{Caruana1997} to more complex recent architectures that utilize both recurrent and convolutional network components 
\citep{Jin2020}. 
Some architectures, such as Cross-stitch networks, use a deep neural network for each task and are not designed to scale to numerous tasks \citep{Misra2016}. 
On the opposite side, context-sensitive networks use a task encoding as input to a network with all parameters being shared\citep{Silver2008}.
A related approach is context adaptation, in which context parameters are learned and used as inputs to a shared neural network\citep{zintgraf19a}.
While much work has centered around neural networks, 
other learners such as support vector machines \citep{Mei2020, Mei2020a}
and Gaussian process regression \citep{Zhou2021} have also been successfully explored.

Even though knowledge sharing has been successful in many cases, it is not guaranteed that all tasks will benefit from each other \citep{standley20a}. 
Negative transfer refers to the phenomenon where the performance of one task is reduced when another task is introduced. 
Deciding which task that should be learning together, and how to best avoid negative transfer, is still an open problem.

We present a multi-task learning based data-driven VFM. 
Our key insight is that knowledge can be shared among wells in a data-driven model, similarly to how knowledge is encoded and reused in mechanistic models. 
In the context of VFMs, we consider modeling the flow rate from one well as a learning task. 
Task domains have different data distributions (domain shift) and the tasks must learn different discriminative models. 
Our MTL architecture, which resembles that of \citep{zintgraf19a}, is specialized for data-driven VFM, for which there is a large number of tasks with few observations. 
It utilizes well-specific parameters to adapt the domains and tasks. 
Domain adaptation is performed by learning domain-specific feature mappings, which transform input features to abstracted \emph{domain features}. 
Task adaptation is enabled by learning \emph{task-specific parameters}. 
The domain features and task parameters are fed to a shared discriminator, to predict flow rates. 
Because our architecture efficiently scales to many tasks, all wells can be modeled simultaneously.

The framing of data-driven VFM as an MTL problem enables us to learn from more data.
While previous methods are limited to small datasets with observations from individual wells, our method scales learning to datasets with observations from any number of wells. 
To test the proposed method, we perform a study of 55 wells from four assets.

\section{Problem description} \label{sec:problem-description}
\begin{figure}[ht]
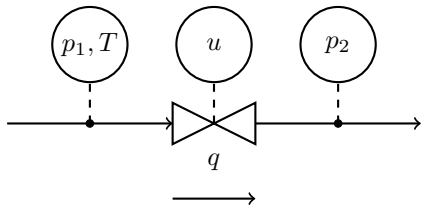

    \centering
    \figwelltopology
    \caption{Choke valve with instrumentation.}
    \label{fig:well-topology}
\end{figure}

The system of interest is the well choke valve. Choke valves are adjustable restrictions that are used to control the flow rate from the well. We only consider measurements that are commonly available for oil and gas wells. These are the pressure ($p_1$) and temperature ($T$) upstream the choke, the pressure downstream the choke ($p_2$), and the choke opening ($u$). In addition, flow rates ($q$) are measured by a separator (well testing) or an MPFM. A single choke valve is illustrated in Figure \ref{fig:well-topology}.

Flow rates are represented as a vector $q^T = \begin{bmatrix} q_{G}, q_{O}, q_{W} \end{bmatrix}$ of gas, oil, and water rate. The rates are customarily given in volumetric flow pr. day, at standard conditions \citep{aime1984si}. However, due to the large magnitude of volumetric gas rates, $q_G$ is scaled down by a factor of 1000 to represent liquid equivalents. We denote the total flow rate by $Q = q_{G} + q_{O} + q_{W}$, and the flow composition fractions by $\phi = q/Q$. 
Flow composition $\phi$ is dependent on reservoir conditions and is slowly time varying. It can be estimated or assumed fixed between well tests.
Here we consider $\phi$ to be known.

We consider the problem of modelling $Q$ given $u$, $p_1$, $p_2$, $T$, and $\phi$. The gas, oil, and water flow rates are then found as $q = Q\phi$. 

\subsection{Insights from mechanistic modelling} \label{sec:mechanistic-model}
The system in question poses some challenges that are best explored by a simple mechanistic example.
For single phase flow, an analytic model
\begin{align}
 Q = A C \sqrt{\frac{p_1 - p_2}{\rho}} , \label{eq:choke-single-phase}
\end{align}
can be derived from the Bernoulli equation \citep{white2008fluid}. 
Here, $A$ is the choke opening area, $C$ is a choke specific flow factor, and $\rho$ is the fluid density. 
Equation \ref{eq:choke-single-phase} is the result of generic assumptions and simplifications, and appears in multiple domains.
Multiphase extensions to Equation \ref{eq:choke-single-phase} are usually domain specific.
Multiplier models are one class of such extensions for oil and gas flows.
They introduce additional factors to Equation \ref{eq:choke-single-phase} to correct errors in the pressure drop calculation due to multiphase flow.
Additionally, the single phase density is replaced by a mixture density. 
There are several variations of multiplier and density computations, some of which are explored in \citep{Schuller2003}. 
These computations often rely on flow composition and fluid properties such as single phase densities. 

Equation \ref{eq:choke-single-phase} contains choke area $A$ as one of the observed variables, and flow factor $C$ as a given constant.
However, these quantities are rarely measured directly.
In the measurement setup considered here, the choke position is given in percent of full travel.
Choke position is not directly comparable between wells, because they have different choke valve designs.
It is common to describe choke valves by a CV curve \citep{Grace2011}.
The CV curve is a mapping between a choke opening and a flow factor, which captures the effect opening area and geometry have on an idealized flow rate.
All mechanistic simulators include CV curves or similar mappings.
Data-driven approaches often circumvent this by modeling directly on $u$, which means the mapping is implicit within a black box model.

To utilize a shared discriminator, it is necessary to adapt the observed values to universally comparable quantities and to capture the unique aspects of each well, such as fluid properties and choke geometries.

\section{Data}
Our data is a set of observations $(Q_{ij}, x_{ij}), i = 1, \dots N_j$, $j = 1, \dots, J$.
Each data point is one observation from one well, indexed as data point $i$ from well $j$.
The total flow rate $Q_{ij}$ is a scalar.
Variables $x_{ij}$ is a vector,
\begin{align}
x_{ij}^\top = \begin{bmatrix} u_{ij}, p_{ij,1}, p_{ij,2}, T_{ij}, \phi_{ij, G}, \phi_{ij, O}, \phi_{ij, W} \end{bmatrix}, \label{eq:explanatory-variables}
\end{align}
of choke opening, pressure upstream choke, pressure downstream choke, temperature, and flow composition fractions. 
Observations are taken at time $t_{ij}$, given in days since the first observation for each well.
Time is used for visualization and splitting datasets.
We are interested in the steady state behaviour of the flow rates.
All observation are therefore averages taken over 3-9 hour intervals of stable production \citep{Grimstad2016}.
Observations are shifted and scaled to lie approximately in the unit interval before model training and evaluation.

\subsection{Data exploration}\label{sec:data-exploration}
The nature of the underlying process and operational practice can create datasets that are challenging for machine learning models to deal with.
For instance, it is common to see reservoir pressure decline as a well develops.
As pressure declines, operators will increase the choke opening to keep flow rates stable at a given target. 
Some assets attempt to reduce the decline, for instance by injecting water into the reservoir \citep{Sheng2014}.
Another remedy is to inject gas into the well flow, which makes the flow composition lighter\citep{guet2006}.
Either way, future operating points will generally not be drawn from the same distribution as the training data. 

\begin{figure*}
    \centering
     \includegraphics[width=\columnwidth, page=1]{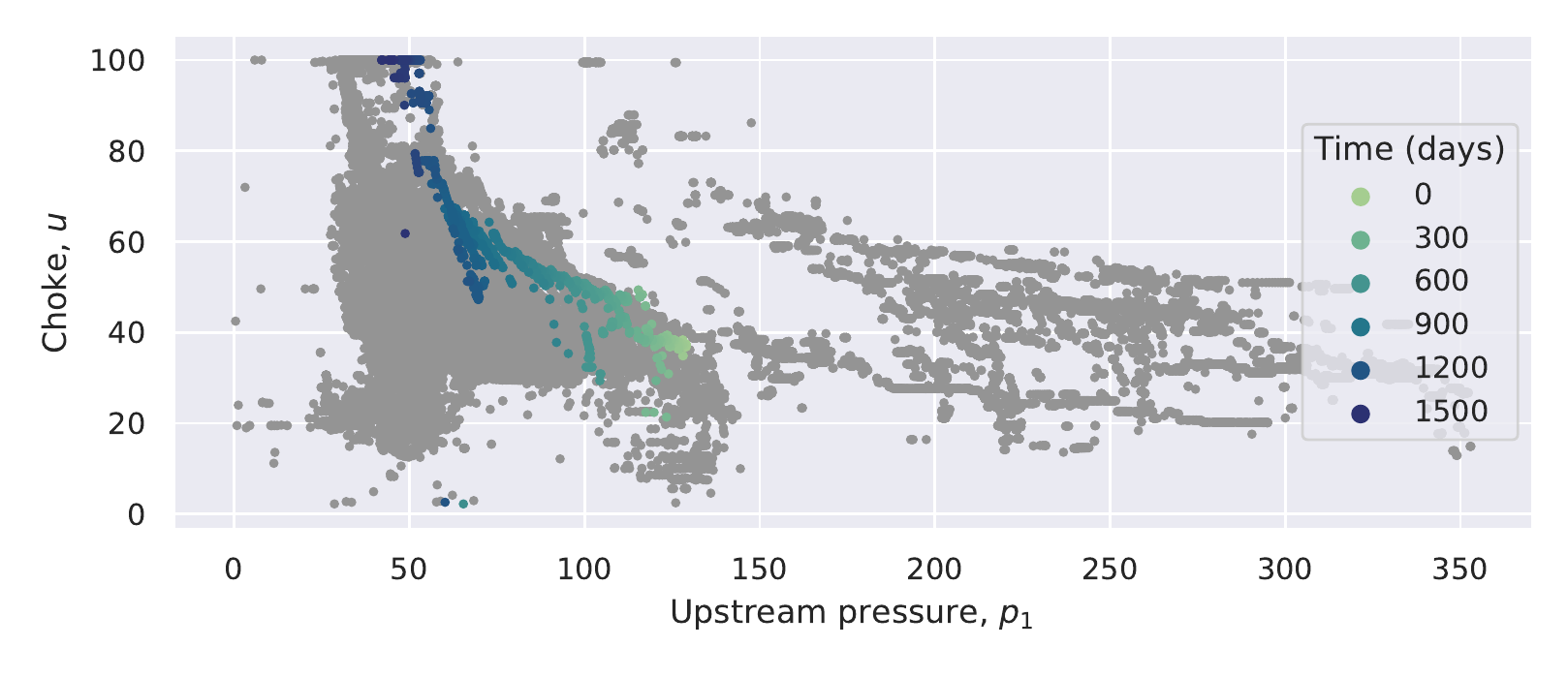} 
    \caption{
    Scatter plot of choke opening and upstream pressure for all wells.
    Observations from a single well is highlighted and coloured by days since first observation. 
    Choke is continuously adjusted to counteract the declining reservoir pressure.
    }
    \label{fig:data-pvu}
\end{figure*}

Figure \ref{fig:data-pvu} illustrates the relationship between choke and pressure from all wells, with one well highlighted and colored by time.
The systematic development in pressure and operational practice is clear. 
Models trained on such data are vulnerable to changes in operational practice.
A similar pattern can be found in the flow composition, which typically develops into a higher water content with time.

All models trained on data from a single well are vulnerable to correlated explanatory variables and how data change with time.
Training on data from multiple wells is one way to overcome these issues. 
A joint data set has several benefits.
The dependencies between explanatory variables become weaker, and the variability within each explanatory variable becomes greater.
This is because different wells have different operating regions and operating patterns.
The reservoir development also becomes less important.
Because, while a single well may move away from its previous operation region, other wells have likely operated under similar conditions before. 

\begin{figure*}
    \centering
     \includegraphics[width=\textwidth, page=1]{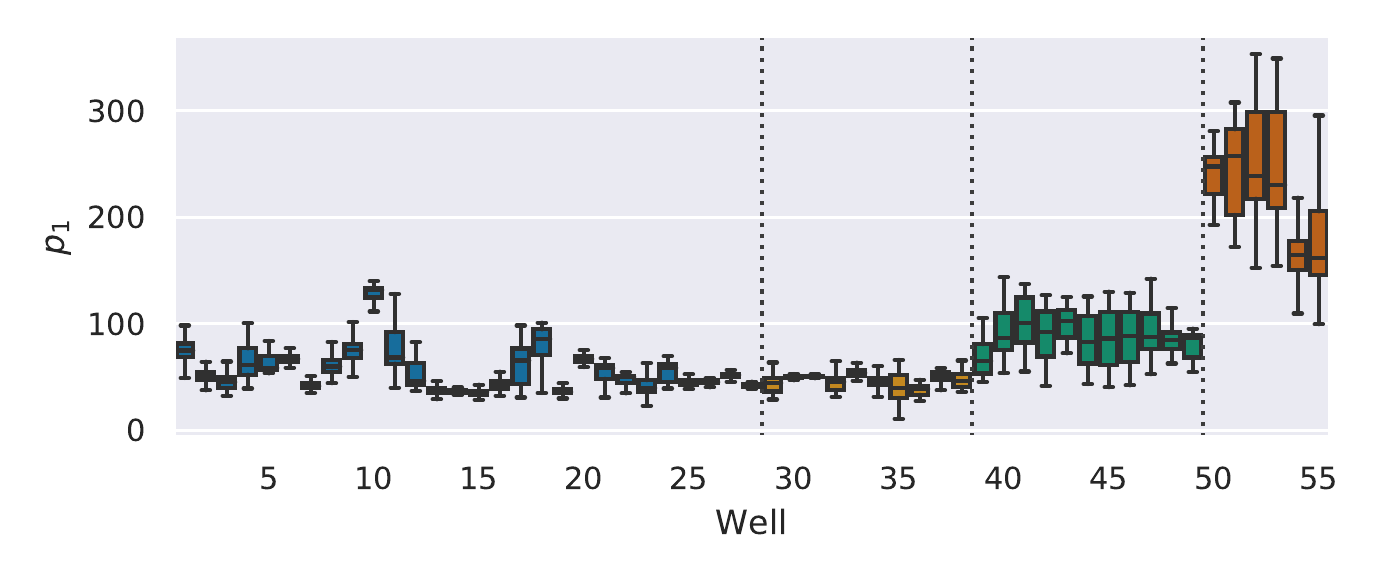} 
    \caption{Box plot of pressure upstream observations for each well.
    The dotted vertical lines and coloring indicate which wells are from the same asset.}
    \label{fig:data-box-p}
\end{figure*}

The joint data set contains data from 55 wells from four assets. 
These wells differ in design, operational practice, and reservoir conditions.
Figure \ref{fig:data-box-p} explore how the distribution of upstream pressure vary between wells and assets.
Many wells have observations in the same range, but one asset is operating at a significantly higher pressure. 

\section{Model formulation} \label{sec:model-formulation}
\begin{figure}
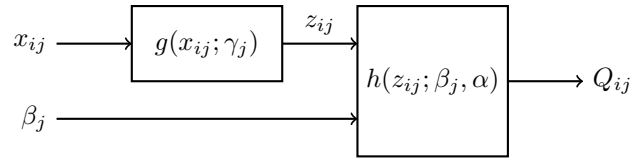

    \centering
    \figarchitecture
    \caption{Block diagram of the model architecture. 
    The model is composed of two functions.
    A task specific domain adaptation $g$, and a flow computation $h$, which takes both task parameters and shared parameters. }
    \label{fig:architecture}
\end{figure}

We propose a data driven virtual flow meter with signature
\begin{align}
	Q_{ij} &= f(x_{ij}; \gamma_{j},  \beta_{j}, \alpha). \label{eq:flow-f}
\end{align}
It takes input variables $x_{ij}$, as described in Equation \ref{eq:explanatory-variables}, 
and is parameterized by three sets of parameters. 
Two of these parameter sets, $\gamma_{j}$ and $\beta_{j}$, are \emph{well specific}, while $\alpha$ is \emph{shared between all wells}.
The model is based on a shared neural network.
The well specific parameters are used in feature adaptation and task differentiation.
The model in Equation \ref{eq:flow-f} is composed of two steps.
A feature adjustment step
\begin{align}
    z_{ij} &= g(x_{ij}; \gamma_{j}), \label{eq:flow-g}
\end{align}
and a flow computation step
\begin{align}
	Q_{ij} &= h(z_{ij}; \beta_{j}, \alpha). \label{eq:flow-h}
\end{align}
The composition is illustrated in Figure \ref{fig:architecture}.

\subsection{Feature adjustment}
As discussed in Section \ref{sec:mechanistic-model}, the observed choke opening is not directly comparable between wells. 
We are interested in a mapping from $u$ for a universally comparable quantity $\psi$,
which is analog to a CV curve.
A piecewise linear mapping is chosen for this purpose.
It has with $m_g$ break points, $u_1^*, \dots, u_{m_g}^*$,
and is parametrized by  $\gamma_{j} = \begin{bmatrix} \gamma_{j,0}, \dots, \gamma_{j,m_g} \end{bmatrix}$.
It is formulated as 
\begin{align}
\psi_{ij} &= (1 + \gamma_{j,0})\left(u_{ij} +
\sum_{k=1}^{m_g} \gamma_{j,k} \max(0, u_{ij}-u_{k}^*)
\right),
\end{align}
which becomes an identity mapping if all parameters are zero. 
A monotonic mapping can be enforced by restricting $\gamma_{j}$, but this is not done here.
In the examples below we use $m_g = 4$ and set breakpoints to $u_{k}^* = 0.2k$. 
Recall that $u_{ij}$ is mapped to the unit interval.

The adjusted feature vector $z_{ij}^\top = \begin{bmatrix}  \psi_{ij}, p_{ij,1}, p_{ij,2}, T_{ij},  \phi_{ij, G}, \phi_{ij, O} \end{bmatrix}$
is then used to evaluate the flow computation.
Note that only two of the flow composition fractions are included.
This is because the fractions sum to one, and the last component is therefore redundant.

\subsection{Flow computation}\label{sec:flow-computation}
\begin{figure}
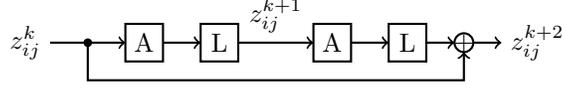

    \centering
    \figrnn
    \caption{Diagram of a neural network residual block as described in Equation \ref{eq:h-residual}. 
    The skip connection span two sets of activation (A) and linear layers (L).}
    \label{fig:residual-block}
\end{figure}

The flow rate approximation $h$ in Equation \ref{eq:flow-h} is modelled by a residual feed forward network. 
The skip connection of the residual blocks spans two hidden layers with pre-activation \citep{He2016}.
There are $m_l$ layers, and all hidden layers have dimension $m_h$.
Linear transforms are parameterized by weights $\alpha = \left\lbrace (W_{k}, b_{k}) | k = 1, \dots, m_l \right\rbrace$.
The rectifier function 
$\Phi(z_{ij}^{k}) = \max(0, z_{ij}^{k})$, where the max operation is performed elementwise,
is used for activation \citep{Goodfellow-et-al-2016}. 
There is no activation on the final layer.
Adjusted features $z_{ij}$ and task parameters $\beta_{j}$ are stacked in a vector before the network is evaluated:
\begin{align}
	z_{ij}^{1} &= \begin{bmatrix} z_{ij} \\  \beta_{j} \end{bmatrix},\\
	z_{ij}^{2} &= W_1 z_{ij}^{1} + b_1, \label{eq:h-first-layer}\\
	\begin{split}
	z_{ij}^{k+2} &= z_{ij}^{k} + W_{k+1} \Phi\left(\left[ W_{k} \Phi\left(z_{ij}^{k}\right) + b_{k}\right]\right) + b_{k+1}, \\
	&k = 2, 4, \dots, m_l-2, \label{eq:h-residual}
	\end{split} \\
	Q_{ij} &= W_{m_l} z_{ij}^{m_l} + b_{m_l}. \label{eq:h-output-regression}
\end{align}
The residual blocks in the neural network is illustrated in Figure \ref{fig:residual-block}.

\subsection{Model comparison}\label{sec:model-types}
The effect of multi-task learning is explored by comparing four different models on the given data.
Two conventional single-task learning models are used as a baseline.
These are compared to two versions of the proposed multi-task architecture.

Both multi-task model formulations are identical, as described above, but they differ in how many tasks are included.  
The first option is trained on wells from the same asset. There are four such models because the dataset contains wells from four assets. 
These are referred to as "MTL-Asset" models.
The second option is trained on all wells and is referred to as the "MTL-Universal" model.
The two multi-task alternatives are selected to explore the degree of positive and negative transfer between tasks. 
These models are collectively referred to as the MTL models.

Gradient boosted trees and conventional neural networks are selected as the single-task baselines, as these represent the current state of the art. 
They are referred to as "STL-GBT" and "STL-ANN" respectively.
Gradient boosted trees are based on the description given in \citep{Bikmukhametov2019}. The neural network models are based on the residual architecture described in Section \ref{sec:flow-computation}, but without the task parameters.
Both baseline models take all observations except water fraction as input, and total flow as output. 
An individual copy of each baseline model is identified for each well. These models are only trained and evaluated on data from a single well.

\section{Method}
Parameters and hyperparameters are found through experimentation and optimization. 
The dataset is divided into development and test sets.
Development data is used to identify hyperparameters and train a final set of models.
Test data is only used to evaluate the performance of the final models.

\subsection{Data splits}\label{sec:data-split}
\begin{figure}
    \centering
    \includegraphics[width=\textwidth]{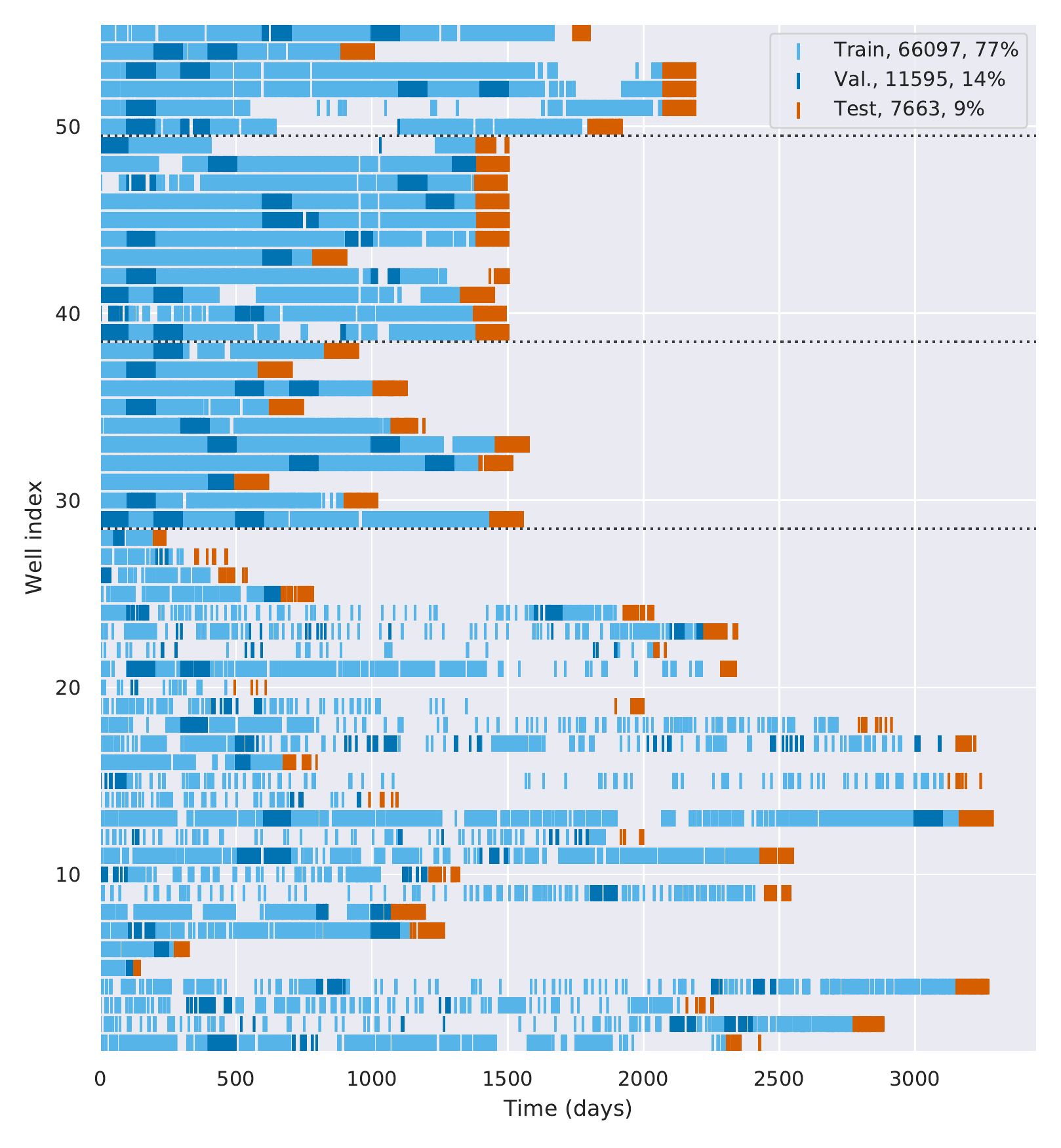} 
    \caption{Training, validation, and test data split for each well. 
    Each mark is one data point.
    For some wells, the time between observations can be significant.
    This can be due to long periods with missing measurements, or because the well was closed.
    Wells from the same asset are grouped by the dotted lines.
    }
    \label{fig:data-split}
\end{figure}

We split the data into subsets used for model development and testing. 
The development dataset is split further into training and validation sets.
Data splits are visualized in Figure \ref{fig:data-split}.

Test data is selected to reflect how models are used in practice.
Meaning that they are trained on all values observed up to a certain point, 
and then used for weeks or months before they are updated again.
For each well, test data is selected such that it comes after development data in time,
and the maximum distance between test and development is 120 days.
The number of points selected is less than 20\% of the observation for that well, and less than 500.

Development data is split further into training and validation. 
Data points are partitioned into blocks of up to 100 consecutive days.
Blocks are randomly divided between training and validation, such that the validation set is 10-20\% of the total. 

\subsection{Loss and minimization} \label{sec:loss}
Model parameters are found by minimizing a standard loss function of prediction error and parameter regularization \citep{hastie2009elements}.
All four model types use weighted mean square error with weights $w_{ij}$ as prediction loss.

For the three neural network models (STL-ANN, MTL-Asset, MTL-Universal) the network parameters are regularized by a $L_2$ norm scaled by a factor $\lambda$. We regularize all parameters except the first neural network bias term.
For the two MTL models, the task specific parameters are regularized by a $L_2$ norm scaled by a factor $\lambda_T$.

The loss is minimized with the AdamW optimizer \citep{Kingma2015}.
The learning rate is set to $10^{-3}$, with a decay rate of $0.5$ every 100 of the last 500 epochs.
Each optimization runs for 3000 epochs during hyperparameter searches. An additional 1000 epochs are used in the final training. There are three batches pr epoch.
Implementation and training are done with PyTorch \citep{Paszke2019}.

STL-GBT models are regularized by penalizing the number of leaves and the squared leaf weight values \citep{Bikmukhametov2019}. 
Implementation and training are done with XGBoost \citep{chen2016}.

\subsection{Model evaluation metrics} \label{sec:metrics}
We use absolute percentage error as the primary performance metric. 
For data point $ij$ with observed flow rate $Q_{ij}$ and predicted flow rate $\hat{Q}_{ij,M}$ from model $M$,
we find percentage error as $e_{ij,M} = 100(\hat{Q}_{ij,M} - Q_{ij})/Q_{ij}$.
For all observations we have $Q_{ij} > 0$.
Model subscripts $M$ indicate which of the four model types the error relates to, e.g., MTL-Asset.
Root mean squared error is used as a secondary metric.

The mean absolute percentage error (MAPE) for well $j$ with model $M$ is denoted by $E_{j,M}$.
Because of the heavy tails of the error distributions,
we use a trimmed mean where 5\% of the largest errors are removed when computing average errors for individual wells \citep{wilcox2010fundamentals}.

In addition to the test set performance, we will explore how the models adhere to the expected physical behavior.
We expect an isolated increase in upstream pressure to increase flow rate,
as indicated by Equation \ref{eq:choke-single-phase}.
For any datapoint $x_{ij}$ we have model predictions $\hat{Q}_{ij,M}$. 
This is compared to $\hat{Q}_{ij,M}^{+}$, which is the same model evaluated on the same data point, 
with the exception that $p_{ij,1}$ is increased by $10$ bar.
We compute a binary score
\begin{align}
    s_{ij,M} =
    \begin{cases} 
        0 & \text{if } \hat{Q}_{ij,M}^{+} - \hat{Q}_{ij,M} > 0, \\
        1 & \text{otherwise},
    \end{cases}
\end{align}
to indicate whether this significant increase in pressure also produces an increase in flow rate.
The average well score is found as $S_{j,M} = \frac{1}{N_j}\sum_{i=1}^{N_j} s_{ij,M}$. A perfect score, $S_{j,M} = 0$, corresponds to a correct sensitivity to changes in upstream pressure for all data points.

\subsection{Hyperparameter selection} \label{sec:result-hps}
Hyperparameters related to model complexity and regularization are optimized for each model individually. E.g., for the STL-ANN models we conduct 55 individual searches. 
Optimization is done by grid search\citep{Bergstra2012}. In case multiple configurations have similar performance, the one with fewer task or network parameters is preferred.

Neural network models are controlled by the number of hidden layers $m_l$, hidden layer dimension $m_h$, and regularization factor $\lambda$.
Additionally, MTL models require task parameter dimension $m_\beta$ and task parameter regularization factor $\lambda_T$.
These parameters are found by grid search,
where candidate values are restricted based on the number of data points available for each model type.

STL-GBT models are tuned by the number of leaves, leaf weight, and the number of boosting iterations. 
These are all found by grid search. 

Sample weight $w_{ij}$ is set to $0.1$ for multiphase meter observations and $1$ for separator observations. 
This is motivated by the high uncertainty in multiphase meters, as discussed in Section \ref{sec:introduction}. 
Since most of the data is from multiphase meters, the results are not particularly sensitive to these values.
These weights are used for all four model types.

\section{Results and discussion} \label{sec:results}

\subsection{Test error overview} \label{sec:result-overview}
We first explore how the models generalize by looking at prediction errors across all wells.
The results are summarized in Table \ref{tab:error-overview}.
The performance is quite similar for the three neural network models,
but multi-task models are more robust towards large errors. 
The neural network models outperform STL-GBT.
The distribution of prediction errors is heavy tailed, with a few outliers skewing the mean errors. 
These outliers motivate the use of trimmed mean when results are reported on a well by well basis.

\begin{table*}
\centering
\begin{tabular}{lrrrrrr}
\hline
Model           &Mean   & P05   & P25   & P50   & P75   & P95   \\
\hline
STL-GBT         & 17.8  & 0.5   & 2.9   & 7.5   & 16.3  & 62.3  \\
STL-ANN         & 20.6  & 0.4   & 2.0   &  4.5  & 11.2  & 44.8 \\
MTL-Asset       & 10.5  & 0.4   & 1.8   &  4.2  &  9.6  & 42.1 \\
MTL-Universal   & 12.8  & 0.5   & 2.0   &  4.4  &  8.6  & 33.1 \\
\hline
\end{tabular}
\caption{Summary statistics of absolute percentage error, $|e_{ij,M}|$.
Statistics are computed on all test data from all wells.
Reported is the mean and a set of percentiles.
}
\label{tab:error-overview}
\end{table*}

\begin{figure}
    \centering
    \includegraphics[width=\textwidth]{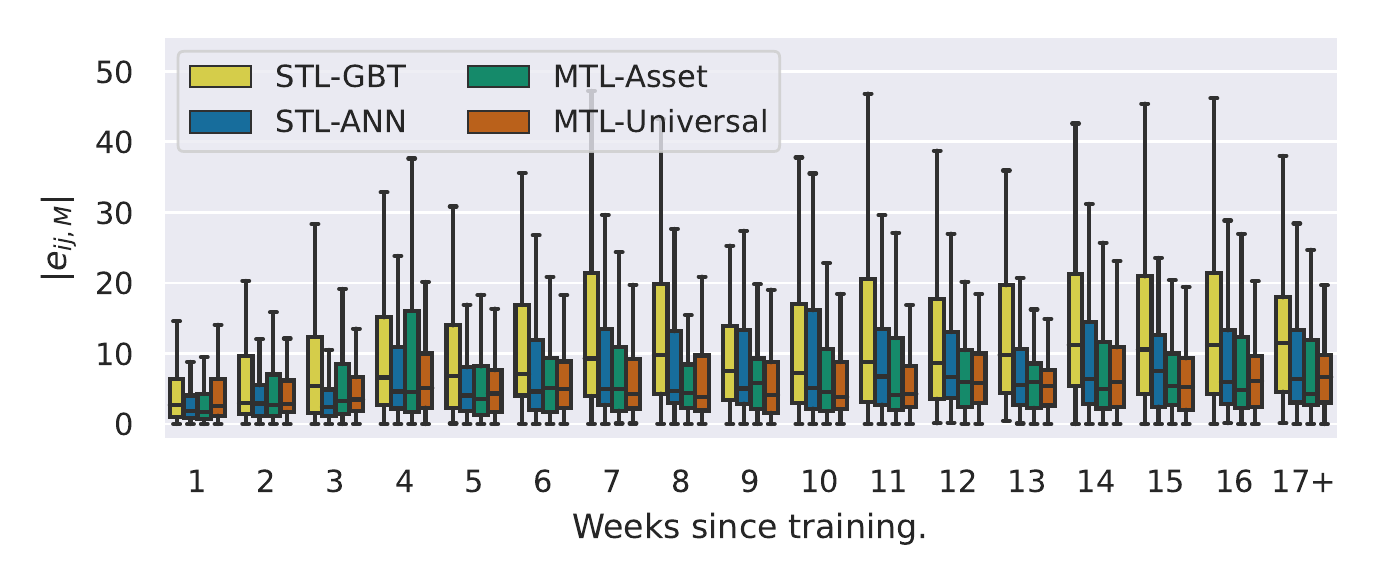} 
    \caption{Box plot of absolute percentage error, grouped by weeks since the last training datapoint. Errors are computed on all test data for all wells.
    }
    \label{fig:test-error-vs-time}
\end{figure}

Figure \ref{fig:test-error-vs-time} illustrates how prediction errors develop with time.
As expected, the performance degrades with time for all model types. 
All model types have great performance in the first few weeks. The benefit of multi-task learning becomes apparent after six weeks.

\subsection{Well by well performance}
Wells have a different number of data points, and the errors reported in Section \ref{sec:result-overview} will naturally be dominated by the wells with many data points.
We now explore the test set errors for individual wells.
Trimmed MAPE and RMSE values for each well is summarized in Table \ref{tab:error-well-by-well}.
The three neural network models have similar performance for the best half of the wells, with performance comparable to conventional multiphase meters.
Multi-task models are significantly better on the more challenging wells.
Neural network models generally outperform STL-GBT.
MAPE and RMSE reveal similar patterns. The remainder of the analysis will focus on MAPE values.

\begin{table}
\centering
\begin{tabular}{llrrrrrr}
\hline
Metric                              & Model         &  Mean &   P05     & P25   & P50   & P75   & P95   \\
\hline
\multirow{4}{*}{$E_{j,M}^{MAPE}$}   & STL-GBT       &  14.5  &    2.3  &    5.8  &    8.6  &   10.8  &   53.9 \\
                                    & STL-ANN       &  10.4  &    1.4  &    3.8  &    5.7  &   11.1  &   34.5 \\
                                    & MTL-Asset     &   8.2  &    1.4  &    3.5  &    6.2  &    9.0  &   22.2 \\
                                    & MTL-Universal &   7.5  &    1.6  &    3.5  &    5.0  &    9.2  &   19.8 \\
\hline
\multirow{4}{*}{$E_{j,M}^{RMSE}$}   & STL-GBT       & 9.6  &    2.0  &    3.7  &    6.3  &   11.0  &   27.0 \\
                                    & STL-ANN       & 7.2  &    1.2  &    2.5  &    4.1  &    8.9  &   22.4 \\
                                    & MTL-Asset     & 5.7  &    1.2  &    2.5  &    4.1  &    7.5  &   14.1 \\
                                    & MTL-Universal & 5.5  &    0.8  &    2.5  &    3.7  &    6.9  &   17.1 \\
\hline
\end{tabular}
\caption{
Mean prediction errors are computed as trimmed MAPE and RMSE for each well. 
This yields 55 error estimates for each combination of model and metric,
which are summarized by their mean and a set of percentiles. 
}
\label{tab:error-well-by-well}
\end{table}

\subsection{Asset performance}
The two MTL models are trained on wells from the same asset and on all wells. 
To explore the degree of positive and negative knowledge transfer, performance is explored on the four assets.
The results are summarized in in Table \ref{tab:error-by-asset}. 
Apart from Asset 3, which has great performance for all neural network models, there is a clear benefit of shared data. 
For assets 1, 2, and 4, the best multi-task model offers a 25-50\% error reduction compared to STL-ANN. 
However, it is an open question to decide which level of data sharing is best suited for a given well or asset. 
All model types struggle with Asset 2,
which could be due to the limited excitation seen in Figure \ref{fig:data-box-p}.
Apart from Asset 2, MTL model performance is close to that expected from conventional multiphase meters.

\begin{table}
\centering
\begin{tabular}{lrrrr}
\hline
Model           & A. 1          & A. 2          & A. 3          & A. 4 \\
STL-GBT         &   15.6        & 13.5          & 10.4          & 18.3 \\
STL-ANN         &   10.9        & 13.8          &  5.9          & 10.5 \\
MTL-Asset       &   8.1         & \textbf{10.2} &  6.5          & 7.9 \\
MTL-Universal   &\textbf{7.3}   & 11.3          &\textbf{5.7}   &\textbf{4.9} \\
\hline
\end{tabular}
\caption{
Model performance grouped by assets.
Reported is the average trimmed MAPE for wells from the same asset, for the four model types. The best model type is highlighted.
}
\label{tab:error-by-asset}
\end{table}

\subsection{Sensitivity analysis}\label{sec:model-sensitivity}
\begin{figure*}
    \centering
    \includegraphics[width=\textwidth]{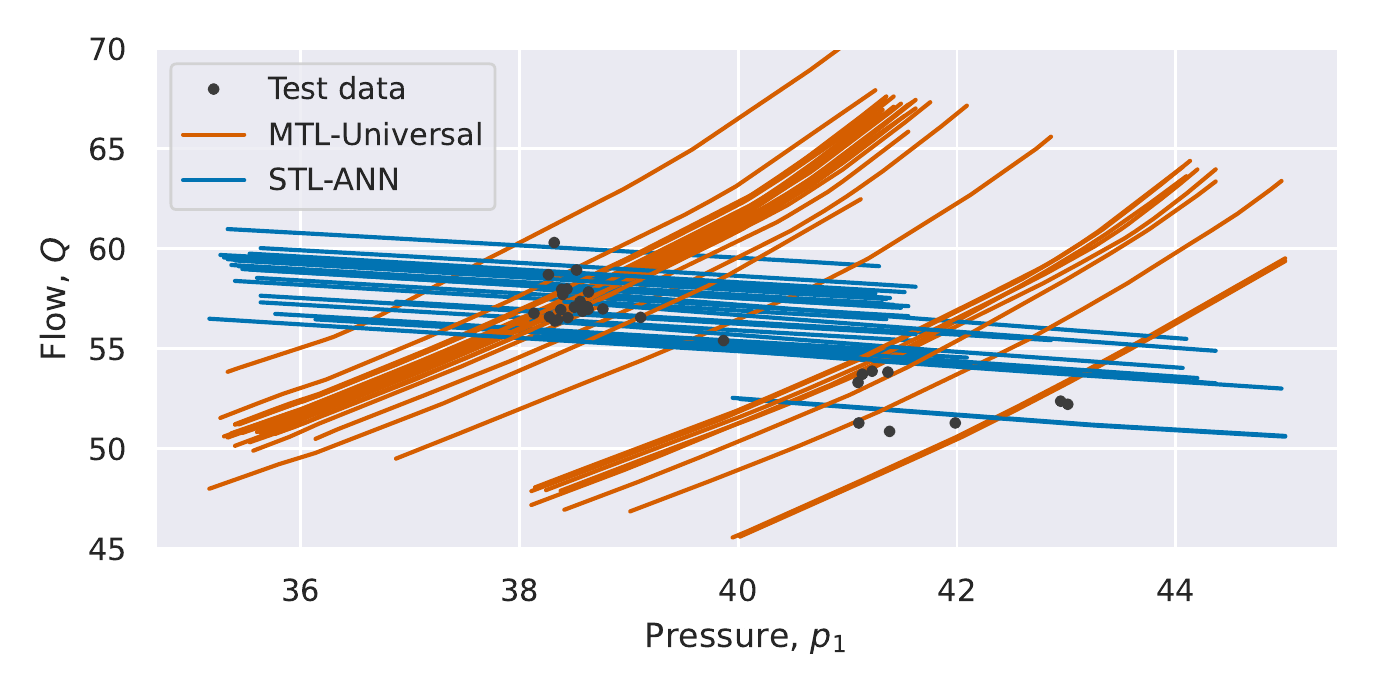} 
    \caption{
    Comparison of model sensitivity for well 7.
    Each model is evaluated by taking a subset of 30 test data points (black dots) and varying upstream pressure in a neighborhood around the observed value.
    The response of STL-ANN is given in blue and MTL-Universal in orange.
    }
    \label{fig:sensitivity}
\end{figure*}

Models with great test set performance can still suffer from the data challenges presented in Section \ref{sec:data-exploration}.
Correlated explanatory variables make it difficult to isolate the effect of individual variables.
Figure \ref{fig:sensitivity} illustrates the issue for Well 7. 
We expect the response to an increase in upstream pressure to be an increase in flow rate,
as indicated by the mechanistic model in Equation \ref{eq:choke-single-phase}.
For well 7, both STL-ANN and MTL-Universal models have low test errors,
with trimmed MAPE being $2.2\%$ and $1.6\%$ respectively.
There is however a significant difference in how they have interpreted the explanatory variables.
In this case, the MTL-Universal model was able to identify the expected response, while the STL-ANN was not.

The observations from Figure \ref{fig:sensitivity} are generalized using
the sensitivity metric described in Section \ref{sec:metrics}.
The results are given in Table \ref{tab:sensitivity}. 
A large majority of wells remain unchanged or achieve a better sensitivity score by sharing data with other wells.
The advantage of transfer learning is clear in this comparison.

\begin{table}[H]
\centering
\begin{tabular}{lr}
\hline
Model           & Error  \\
\hline
STL-GBT         & 0.56  \\
STL-ANN         & 0.28  \\
MTL-Asset       & 0.14  \\
MTL-Universal   & 0.07  \\
\hline
\end{tabular}
\caption{
Mean sensitivity error $S_{j,M}$ for the four model types. 
Zero is the best score, which means a model had the correct sensitivity for each data point for a given well.
}
\label{tab:sensitivity}
\end{table}

\subsection{Ablation study}\label{sec:ablation-study}

An ablation study is conducted to better understand the improvements seen in the proposed architecture \citep{lipton2019}.
The proposed model extends the current state-of-the-art with multi-task learning. Two task adaptation mechanisms are included. These are task parameters $\beta_j$, and domain adaptation parameterized by $\gamma_j$.
To explore the effect of the task adaptation, the MTL-Universal model is trained with one or both elements removed. When both elements are removed, the model is reduced to a single-task neural network trained on data from all wells. New hyperparameters are found for each ablation.
The results is given in Table \ref{tab:ablation}.
On average, both adaptation mechanisms have a similar impact. 
The inclusion of both is beneficial for overall performance, but with diminishing returns, as they are potentially overlapping.

\begin{table}[H]
\centering
\begin{tabular}{lr}
\hline
Ablation                        & Error \\
\hline
Remove $\gamma_j$ and $\beta_j$ & 10.5  \\
Remove $\beta_j$                &  8.8  \\
Remove $\gamma_j$               &  8.4  \\
Complete model                  &  7.5 \\ 
\hline
\end{tabular}
\caption{
Summary of ablations conducted on the MTL-Universal model.
Average trimmed MAPE is computed on all wells.
Value for the complete model is repeated from Table \ref{tab:error-well-by-well}.
}
\label{tab:ablation}
\end{table}

\subsection{Model complexity}
A multi-task model is more complex than an isolated single-task model.
However, as the number of tasks grows, there are several aspects to multi-task learning that leads to overall less complexity.
Table \ref{tab:model-complexity} summarizes the number of parameters and training time for the four model types presented.
STL-GBT is a separate class of models and is much faster to compute than neural networks.
For neural network models, the larger models require more time for each model, but less time overall.

In the universal model, the number of well parameters is significantly smaller than the number of parameters needed in an individual well model.
On average, an MTL-Asset model requires almost the same number of parameters as MTL-Universal.
Indicating that model size does not need to grow significantly with the number of tasks.

\begin{table}[H]
\centering
\begin{tabular}{lrrr}
\hline
Model           & Models    & Time      & Parameters \\
\hline
STL-GBT         & 55        & 00:57     & -         \\
STL-ANN         & 55        & 56:06     & 450455    \\
MTL-Asset       & 4         & 21:25     & 711389    \\
MTL-Universal   & 1         & 12:38     & 203851    \\
\hline
\end{tabular}
\caption{
Summary of model complexity, judged by the number of parameters and time required to train models for all wells.
E.g., it took 21 minutes and 25 seconds to train the four MTL-Asset models, and they have 711389 parameters in total.
All models are trained on a single GPU.
}
\label{tab:model-complexity}
\end{table}

In terms of manual work and maintenance, 
the MTL-Universal formulation scales better with additional wells than conventional model formulations, 
because it is only one neural network that must be curated.
This is highly advantageous for the practical application and commercialization of the results.

\subsection{Qualitative properties}\label{sec:result-explore}
\begin{figure*}
    \centering
    \includegraphics[width=\textwidth]{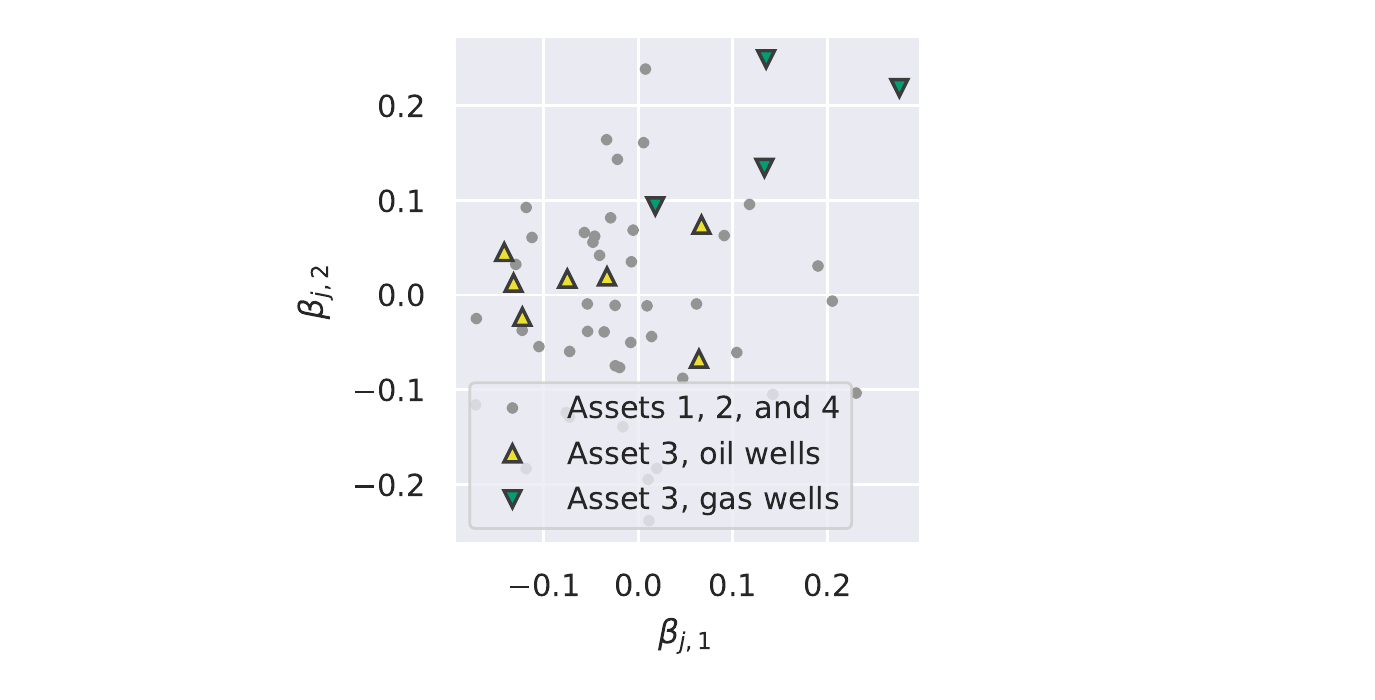} 
    \caption{Plot of $\beta_{j}$ from the universal model. 
    Asset 3 has two distinct classes of wells, oil producers and gas producers, which are highlighted by triangle markers.
    }
    \label{fig:beta-scatter}
\end{figure*}

\begin{figure*}
    \centering
    \includegraphics[width=\textwidth]{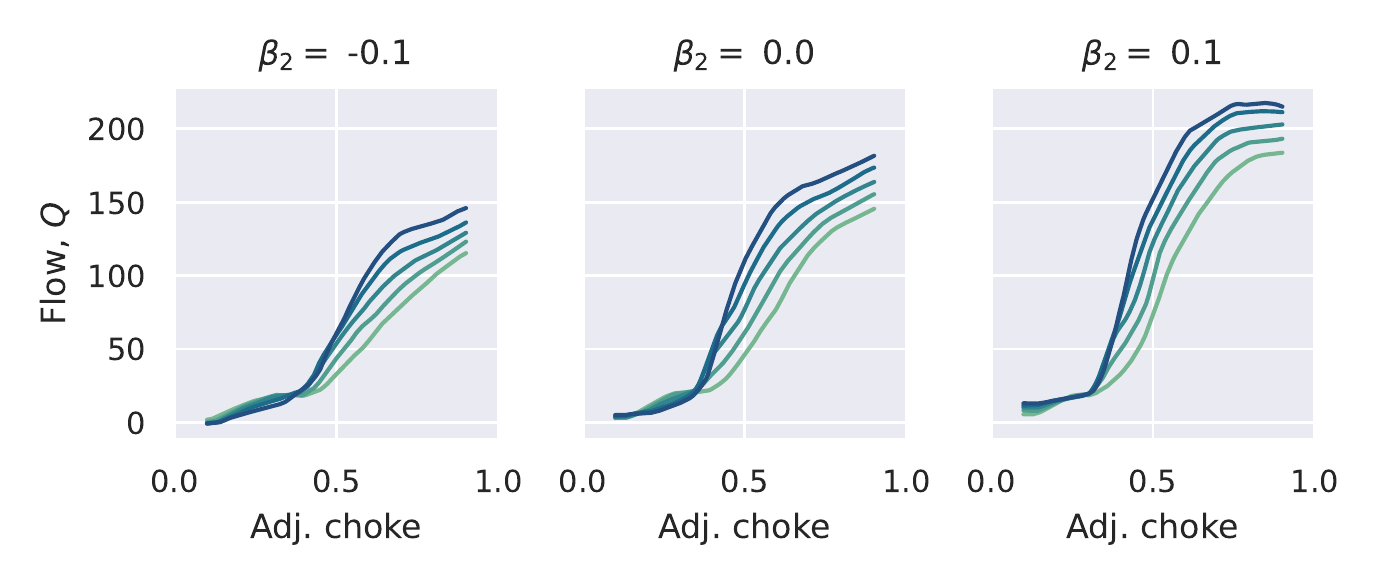} 
    \caption{
    Effect of $\beta_{j}$ on model predictions.
    The universal model is evaluated on a fixed operating point, except for varying the adjusted choke between 0.1 and 0.9. 
    Each subplot has a fixed value for $\beta_{j,2}$, given in the title.
    Each curve has a fixed value for $\beta_{j,1}$, ranging from -0.1 for the light green to 0.1 for the dark blue. 
    }
    \label{fig:beta-curve}
\end{figure*}

The selected hyperparameter configuration has two task parameters for each well,
$\beta_{j}^T = \begin{bmatrix} \beta_{j,1}, \beta_{j,2}, \end{bmatrix}$.
Figure \ref{fig:beta-scatter} illustrates the identified parameters for all wells. 
Asset 3 has two types off wells, oil producers and gas producers, which have different choke geometries and fluid properties. 
These wells are separated in space according to their classes,
which indicates that task parameters capture physical properties, rather than being proxies for the well index.
To further support this,
Figure \ref{fig:beta-curve} illustrates how changes in task parameters alter the shape and magnitude of the model response in a consistent fashion.

\section{Conclusion}
A multi-task learning architecture for data driven virtual flow metering was proposed and explored in a study of 55 wells from four assets.
The proposed architecture successfully addresses the identified data challenges, while generally improving model performance.   
Sharing data between wells improves robustness towards changes in operational practice and makes the model adhere better to the expected physical relationships.
In terms of prediction errors, all assets benefit from some level of data sharing.
Two of the assets saw average errors reduced by 30--50\% when data was shared between all assets.
One asset saw a reduction of 24\% when data was shared within the asset, but only 16\% improvement when data is shared between all assets. This indicates that issues related to negative transfer could be present in the VFM problem.
The final asset saw no significant improvements because the single task architectures already performed well.
Overall, the MTL architecture is a promising step towards a data driven virtual flow meter solution.

\subsection{Future work}
All wells explored here have many data points.
It is expected that wells with fewer observations will see a greater benefit from the knowledge sharing architecture.
Exploring these opportunities is left as future work.
Additionally, it is desirable to further explore task synergies and negative transfer in the VFM context.

In the proposed model, task specific parameters are constants.
In practice, these are likely time varying, since both the well and reservoir will develop over time,
e.g., changes in fluid properties, or equipment wear and tear. 
These aspects are topics for future research.

\section*{Acknowledgements}
This work was supported by Solution Seeker Inc. and The Research Council of Norway.

\bibliography{references}

\end{document}